%% file: nac_nips.tex
\documentclass{article}

\usepackage[preprint]{nips_2018}

\usepackage[utf8]{inputenc} 
\usepackage[T1]{fontenc}    
\usepackage{hyperref}       
\usepackage{url}            
\usepackage{booktabs}       
\usepackage{amsfonts}       
\usepackage{nicefrac}       
\usepackage{microtype}      
\usepackage{amsmath}

\usepackage{subcaption}
\usepackage{tikz}
\usetikzlibrary{shapes,arrows}
\usepackage{verbatim}

\usepackage{array}
\newcolumntype{L}[1]{>{\raggedright\let\newline\\\arraybackslash\hspace{0pt}}m{#1}}
\newcolumntype{C}[1]{>{\centering\let\newline\\\arraybackslash\hspace{0pt}}m{#1}}
\newcolumntype{R}[1]{>{\raggedleft\let\newline\\\arraybackslash\hspace{0pt}}m{#1}}

\usepackage[ruled]{algorithm2e} 

\usepackage{graphicx}
\usepackage{booktabs}
\setcitestyle{numbers, brackets={square}}

\title{Fast Neural Architecture Construction using EnvelopeNets}

\author{
  Purushotham Kamath \\
  Cisco Systems\\
  \texttt{pukamath@cisco.com} \\
  \And
  Abhishek Singh \\
  Cisco Systems \\
  \texttt{abhishs8@cisco.com} \\
  \And
  Debo Dutta \\
  Cisco Systems \\
  \texttt{dedutta@cisco.com}  \\
}

\begin{document}

\maketitle

\begin{abstract}
\input{abstract}
\end{abstract}

\input{intro}
\input{related}
\input{motivation}

\input{hypothesis}

\input{methodology}

\input{results}
\input{discussion}
\input{conclusion}

\bibliographystyle{ACM-Reference-Format}
\bibliography{bib} 
\input{appendix}

\end{document}

%% file: abstract.tex
Fast Neural Architecture Construction (NAC) is a method to construct deep network architectures by pruning and expansion of a base network.  
In recent years, several automated search methods for neural network architectures have been proposed using methods such as evolutionary algorithms and reinforcement learning.
These methods use a single scalar objective function (usually accuracy) that is evaluated after a full training and evaluation cycle.
In contrast NAC directly compares the utility of different filters using statistics derived from {\em filter featuremaps} reach a state where the utility of different filters within a network can be compared and hence can be used to construct networks.
The training epochs needed for filters within a network to reach this state is much less than the training epochs needed for the accuracy of a network to stabilize. 
NAC exploits this finding
to construct convolutional neural nets (CNNs) with close to state of the art accuracy, in < 1 GPU day, faster than most of the current neural architecture search methods.
The constructed networks show close to state of the art performance on the image classification problem on well known datasets (CIFAR-10, ImageNet) and consistently show better performance than hand constructed and randomly generated networks of the same depth, operators and approximately the same number of parameters.



%% file: intro.tex
\section{Introduction}

In recent years, several bespoke neural networks (InceptionNet~\cite{szegedy15}, DenseNet~\cite{huang16}, and ResNet~\cite{he15}) have shown significant improvements on the image classification and object detection problems. 
More recently, search algorithms and recurrent networks have found network architectures that outperform these bespoke architectures.

The time needed to discover a network by these algorithms is fundamentally limited by the need to run a full training and evaluation cycle for every iteration of the construction algorithm.
It is known that some networks converge faster than others depending on the structure, hyperparameters and other factors~\cite{bergstra11}. 
Hence search methods have to wait for accuracy to converge before comparing networks.
The search space for construction of these networks is exponential in the number of operators and each iteration requires waiting for this convergence.
The search algorithms address the long search and evaluation times by various methods including cell search with stacked cell design (rather than a network search)~\cite{zoph17b}, parameters prediction and sharing~\cite{pham18,brock17}, shorter training runs~\cite{zhong18} and other methods.

This work proposes construction methods that are not dependent on running a full training and evaluation cycle.
They are based on intuition from previous work that indicates different stages of the network play different roles in the overall classification task.
Zeiler et. al.~\cite{zeiler13} show that shallower layers of a network extract fine features while deeper layers extract grosser features.
Li et. al.~\cite{haoli16} show that pruning can be effective in reducing the size of a network without significant degradation in accuracy.
Both of these indicate that, at different layers of a network, some filters are more important than others.

This work shows that statistics obtained from the featuremaps at the outputs of the filters of a network during training, {\em i.e. featuremap statistics}, can be used to compare the utility of filters within a network.
These statistics reach a state where the utility of different filters within a network and hence, their relative importance to the classification (or other) task can be evaluated.
The time needed for filters to reach this state, is much less than the time needed for the accuracy of a network to converge {\em i.e.} the time needed for an accurate comparison of the performance of two networks. 
Therefore the pruning and expansion can be done without needing to wait for a full training cycle to complete. 
A filter's utility is calculated based on statistics obtained from their featuremaps obtained during the training.
These statistics reach a state where the utility of different filters within a network 
can be evaluated.
Experimentally, we find that the time needed for filters to reach this state, is much less than the time needed for the accuracy of a network to converge {\em i.e.} the time needed for an accurate comparison of the performance of two networks, resulting in a speedup in the construction time.
{\em EnvelopeNet} construction exploits this property for the iterative construction of convolutional neural networks.

The pruning and expansion algorithm fits in well with intuition from previous work that indicates different layers of the network play different roles in the overall classification task ~\cite{zeiler13, olah2018}. The layers closer to the head of the network extract gross features (edges, boundaries, shapes) while deeper layers compose these into more abstract features (such as facial features). 
Further,~\cite{greff16} indicates that each stage of a network iteratively refines its estimates of the same features.
In InceptionNet~\cite{chollet16}, the 3 parallel paths with different filters were shown to extract features at different levels.
After training, it was found that for most layers, one of the paths dominated the others, indicating that one path was primarily activated at each layer.

The algorithm also mirrors theories on the ontogenesis of neurons in the brain.
Brain development is believed to consist of neurogenesis~\cite{ngenesis_1}, where the neural structure initially develops, gradually followed by apoptosis~\cite{apoptosis}, where neural cells are eliminated, introduction of more neurons by hippocampal neurogenesis~\cite{Eriksson1998} and synaptic pruning~\cite{syn_pruning}, where synapses are eliminated.
The NAC algorithm consists of analogous steps run in iterations: model initialization with a prior (neurogenesis), a truncated training cycle, pruning filters (apoptosis), adding new cells (hippocampal neurogenesis) and pruning of skip connections (synaptic pruning).

%% file: related.tex
\section{Related work}

Recent work in automated neural network architecture design/search can be broadly classified into two categories. 

\begin{itemize}
\item Network design: Networks are designed through combinatorial search, evolutionary algorithms or recurrent networks. {\em E.g.} Neuroevolution~\cite{floreano08}, AmoebaNet~\cite{real18}, Bayesian Optimization~\cite{kandasamy18}, MetaQNN~\cite{baker16}, Genetic CNN~\cite{xie17}
\item Cell or block design: Search, evolutionary, recurrent networks or other methods are used to find a cell (or block) of operators. Multiple cells are stacked in series to form a network. {\em E.g.} NasNet~\cite{zoph17b}, BlockQNN~\cite{zhong18} and Dutta et. al. ~\cite{dutta18}.
\item Optimization based methods: {\em E.g.} DARTs~\cite{darts}, NAO~\cite{nao}
\end{itemize}

Neuroevolution methods~\cite{floreano08} encompass a range of evolutionary algorithms/techniques that discover network architectures. 
Real et. al.~\cite{real17} proposed an evolutionary algorithm to pick a combination of architecture and hyperparameters. 
A regularized version of this technique called AmoebaNet~\cite{real18} improves performance and starts the neuroevolution from a prior.
Elsken et. al.~\cite{elsken17} use hill climbing to incrementally build neural networks. 
Traditionally, hyperparameter tuning has used a variety of blackbox techniques such as grid search, Bayesian optimization and random search~\cite{bergstra12}.
These techniques are effective for continuous valued parameters such as the number of layers and filter sizes, but are hard to apply to network architecture.
Kandasamy et. al.~\cite{kandasamy18} proposes a distance metric to apply Bayesian optimization to the architecture search.
Neural Architecture Search~\cite{zoph17a} uses an RNN that generates the number of filters, filter size, and stride for a convolution network.
Super neural networks use top down approach by designing a large network fabric, based on previously proposed networks, and recovering architectures by selecting a path over an ensemble of architectures. Convolutional Neural fabrics~\cite{saxena16}, PathNet~\cite{pathnet}, Budgeted SuperNets~\cite{veniat17} adopt this strategy.


The large search space of the neuroevolution methods lead to the search for cell architectures. 
Cell based design was influenced by bespoke hand designed networks such as InceptionNet~\cite{szegedy15} and XceptionNet~\cite{chollet16}.
These networks have repeated structures of cells and/or connections and/or repeated design motifs.
NasNet\cite{zoph17b} uses recurrent networks with reinforcement learning to generate and optimize cell designs composed from a fixed set of operators (blocks).
Subsequent work~\cite{zoph17b} showed the transferability of these methods, addressing concerns that construction methods are susceptible to overfitting.
The paper observed that the cells they generate are often envelopes over a broad class of human invented architectures, motivating the use of envelopes in our construction method.
Progressive NAS~\cite{liu17} extends the work by searching for a good cell composed from a limited set of blocks using a sequential model-based optimization strategy. 

The performance of both cell and network based designs have exceeded that of the state of the art bespoke networks.  
However, both the network as well as the cell construction methods need computation resources for both the search phase as well as the evaluation phase. 
Several methods have been proposed to reduce the computation requirements.
Efficient NAS~\cite{pham18} reduces the computation resources needed for NasNet through parameter sharing across iterations of generation and reducing the search space. 
BlockQNN~\cite{zhong18} uses an early stop mechanism to stop training early with reward being function of accuracy at early stop as well as model complexity.
Architecture search by network transformation~\cite{cai17} is another method that uses parameter inheritance along with network transformation.  
DeepArchitect~\cite{negrinho17}  and Liu et. al.~\cite{liu18} use efficient search methods by representing the search space in a structured manner.
SMASH~\cite{brock17} avoids the full training of candidate models by generating their weight from Hypernets~\cite{HaDL} and predicting accuracy. 


Outside the field of architecture search, model reduction methods that analyze filter statistics have been used to reduce model size for inference on resource constrained platforms, without substantial loss of accuracy.
Li et. al.~\cite{haoli16} showed that filter pruning of convolutional networks can reduce parameters, training and inference times without significant degradation in accuracy.
Roy et. al.~\cite{roy18}, and Molchanov et. al.~\cite{molchanov16} also use pruning to remove duplicates and to reduce model size.
Mittal et. al~\cite{mittal18} surveyed several metrics for pruning and showed that random pruning can be as effective as algorithmic pruning when performed on a hand designed network (such as ResNet~\cite{he15}).

Most of the cell and network search methods are bottom up methods, agnostic to filter statistics, while model reduction has traditionally been top down, using filter statistics. 
EnvelopeNets integrates techniques from both domains (evolution and pruning) to enable construction using featuremap statistics.



%% file: motivation.tex
\section{Motivation and Hypothesis}
The motivation of this work is to answer three questions:

\begin{itemize}
\item Do there exist {\em featuremap statistics}, (any statistic extracted from the time series of the featuremaps of the filters in the network during training) that reach a state, {\em featuremap stability}, when the performance of the filters can be compared? 
\item Can networks with improved accuracy be constructed automatically using featuremap statistics to control the construction process? Do the constructed networks perform significantly better than equivalent arbitrarily constructed or randomly generated networks?
\item Do the featuremap statistics reach stability significantly faster than the network accuracy converges {\em i.e.} the time needed to make a reliable comparison between the accuracy of two networks?
\end{itemize}

Our construction method is based on EnvelopeNets. An {\em EnvelopeNet} is a deep convolutional neural network of stacked EnvelopeCells. 
{\em EnvelopeCells} are supersets (or envelopes) of previously proposed handcrafted and/or generated cells constructed from basic {\em operators} or {\em blocks}
The network is structured in  {\em stages}. A stage is a sequence of layers separated by
widening cells, which are layer of cells that increase the channel width of the image.
The construction method iteratively restructures the EnvelopeNet based on the utility of the filters within the network.
Note that the definition of featuremap stability does not imply convergence of the featuremap statistic.

%% file: hypothesis.tex

Our hypothesis, that EnvelopeNets can be restructured to yield higher performing networks, is based on intuition from related work around cell construction, network search and pruning filters and connections.
Part of the intuition behind the algorithm comes from visualization techniques that indicate what the individual layers of a neural network perform~\cite{zeiler13,olah2018}.
The studies indicate that after training a network, the layers closer to the head of the network extract gross features (edges, boundaries, shapes) while deeper layers compose these into more abstract features or objects (such as meshes, facial features). 
Their results also show that visual inspection of filter performance can be used for architecture selection and can have a significant impact on performance. 
The reasoning is that after a reasonable amount of training is complete, filters generally identify the scale of the features which they extract. 
The filters that extract less important features would be better suited to be placed in a layer where they can contribute more to the classification task. 
This hypothesis is supported by the results of Zeiler et. al.~\cite{zeiler13}, that show that, after training, some filters in a deep network end up with featuremaps with "dead" features  while other have cleaner distinctive features.
Further, ~\cite{greff16} indicates that each stage of a network iteratively refines its estimates of the same features. 
This has influenced the design of the EnvelopeNet which is structured in stages, and construction, which is done independently on each stage.



The algorithms described in this work are non optimal.
However, we show empirically, that they generate constructions whose performance exceeds that of the EnvelopeNet, several arbitrarily constructed and randomly generated networks of the same network complexity (same depth, same blocks and approximately the same number of parameters).

%% file: methodology.tex
\section{Construction using EnvelopeNets}
\label{sec:methodology}
\begin{figure*}
\centering
\includegraphics[width=14cm,height=1.5cm]{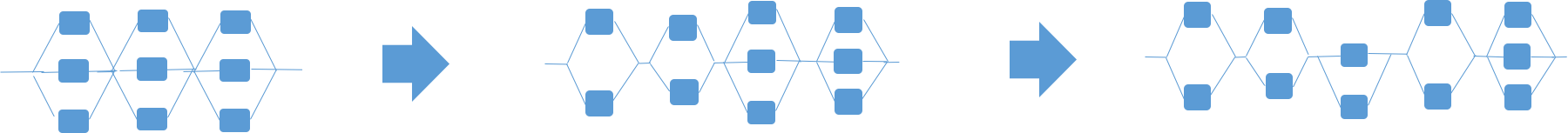}
\caption{Neural Architecture Construction (NAC) using EnvelopeNets}
\label{fig:nac}
\end{figure*}

Construction has three components: the design of the EnvelopeNet, the choice of the featuremap statistic and the construction algorithm. 

\subsection{EnvelopeNet}

The EnvelopeCell is a set of $M$ convolution blocks connected in parallel. {\em E.g.} one of the EnvelopeCells used in this work has 6 convolution blocks connected in parallel: 1x1 convolution, 3x3 convolution, 3x3 separable convolution, 5x5 convolution, 5x5 separable convolution and  7x7 separable convolution. This is a subset of the blocks used in the cell discovered in~\cite{pham18}.
Each block consists of a convolution block, a Relu unit and a batch normalization. 
In addition the network has three additional types of cells: Wideners (a maxpool unit and 3x3 convolution filter connected in parallel) which downsample the image dimensions by a factor of two and double the channel width, a stem (initial cell) for the network that increases the input channel width $C$ for the EnvelopeNet and a classification cell consisting of an average pooling block, a fully connected layer with dropout and softmax.

The EnvelopeNet consists of a number of the EnvelopeCells stacked in series organized into stages of $n_i$ layers, separated by wideners.
The cells are organized in stages, with wideners separating each stage. 
The stem and classification blocks are placed at the head and tail of the network.

We refer to a network using the notation, $C$/$n_1$-$n_2$-$n_3$.../$M$ where $C$ is the number of input channels to the EnvelopeNet (output of the stem cell), $n_i$ is the number of layer in stage $i$ and $M$ is the number of filters in an EnvelopeCell. {\em E.g.} in this work one of the EnvelopeNets we use is an {\em 128/10-1-1-1/6} EnvelopeNet

\subsection{Featuremap statistic}

\tikzstyle{block} = [draw, fill=blue!20, rectangle, 
    minimum height=1.15em, minimum width=1.2em]
\tikzstyle{concat} = [draw, fill=green!20, circle, node distance=0cm]
\tikzstyle{feature_map} = [draw, fill=blue!20, circle, node distance=0.2cm, inner sep=0pt]
\tikzstyle{input} = [coordinate]
\tikzstyle{output} = [coordinate]
\tikzstyle{pinstyle} = [pin edge={to-,thin,black}]
\tikzstyle{arrow} = [thick,->,>=stealth]
\tikzstyle{dotted-arrow} = [thick,dashed,->,>=stealth]

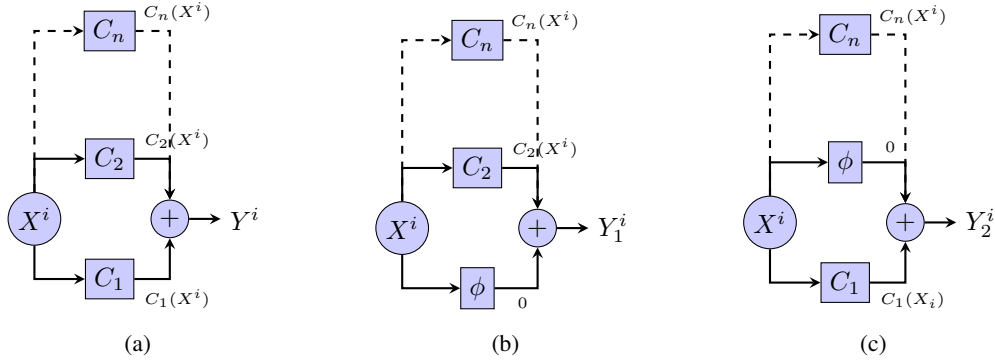
\begin{figure}[h]
\captionsetup[subfigure]{font=footnotesize}
\subcaptionbox{\label{fig:Y}}[0.3\textwidth]{%
    \begin{tikzpicture}
    	\node [input, name=input] {};
        \node [feature_map, minimum size=0.7cm, right of=input] (fmap) {$X^i$};
        \node [block, right of=fmap, yshift = 0.8cm, xshift = 0cm] (W1) {$C_2$};
        \node [block, right of=fmap, yshift = -0.8cm, xshift = 0cm] (W2) {$C_1$};
        \node [block, right of=fmap, yshift = 2.5cm, xshift = 0cm] (W3) {$C_n$};
        \node [feature_map, minimum size=0.5cm, right of=fmap, xshift=1.6cm] (concat_op) {$+$};
        \node [opacity=.0,text opacity=1, right of=concat_op, xshift=0cm] (y) {$Y^i$};
        \draw [arrow] (fmap) |-  (W1);
        \draw [arrow] (fmap) |-  (W2);
        \draw [arrow] (W1) node[above, xshift=0.9cm] {\tiny $C_2(X^i)$} -| (concat_op);
        \draw [arrow] (W2) node[below, xshift=0.9cm] {\tiny $C_1(X^i)$} -| (concat_op);
        \draw [arrow] (concat_op) -- (y);
        \draw [dotted-arrow] (fmap) |- (W3);
        \draw [dotted-arrow] (W3) node[above, xshift=0.9cm] {\tiny $C_n(X^i)$} -| (concat_op);
    \end{tikzpicture}}%
\subcaptionbox{\label{fig:Y1}}[0.4\textwidth]{%
    \begin{tikzpicture}
    	\node [input, name=input] {};
        \node [feature_map, minimum size=0.7cm, right of=input] (fmap) {$X^i$};
        \node [block, right of=fmap, yshift = 0.8cm, xshift = 0cm] (W1) {$C_2$};
        \node [block, right of=fmap, yshift = -0.8cm, xshift = 0cm] (W2) {$ \phi $};
        \node [block, right of=fmap, yshift = 2.5cm, xshift = 0cm] (W3) {$C_n$};
        \node [feature_map, minimum size=0.5cm, right of=fmap, xshift=1.6cm] (concat_op) {$+$};
        \node [opacity=.0,text opacity=1, right of=concat_op, xshift=0cm] (y) {$Y_1^i$};
        \draw [arrow] (fmap) |-  (W1);
        \draw [arrow] (fmap) |-  (W2);
        \draw [arrow] (W1) node[above, xshift=0.9cm] {\tiny $C_2(X^i)$} -| (concat_op);
        \draw [arrow] (W2) node[below, xshift=0.6cm] {\tiny $0$} -| (concat_op);
        \draw [arrow] (concat_op) -- (y);
        \draw [dotted-arrow] (fmap) |- (W3);
        \draw [dotted-arrow] (W3) node[above, xshift=0.9cm] {\tiny $C_n(X^i)$} -| (concat_op);
    \end{tikzpicture}}%
\subcaptionbox{\label{fig:Y2}}[0.3\textwidth]{%
    \begin{tikzpicture}
    	\node [input, name=input] {};
        \node [feature_map, minimum size=0.7cm, right of=input] (fmap) {$X^i$};
        \node [block, right of=fmap, yshift = 0.8cm, xshift = 0cm] (W1) {$\phi$};
        \node [block, right of=fmap, yshift = -0.8cm, xshift = 0cm] (W2) {$C_1$};
        \node [block, right of=fmap, yshift = 2.5cm, xshift = 0cm] (W3) {$C_n$};
        \node [feature_map, minimum size=0.5cm, right of=fmap, xshift=1.6cm] (concat_op) {$+$};
        \node [opacity=.0,text opacity=1, right of=concat_op, xshift=0cm] (y) {$Y_2^i$};
        \draw [arrow] (fmap) |-  (W1);
        \draw [arrow] (fmap) |-  (W2);
        \draw [arrow] (W1) node[above, xshift=0.6cm] {\tiny $0$} -| (concat_op);
        \draw [arrow] (W2) node[below, xshift=0.9cm] {\tiny $C_1(X_i)$} -| (concat_op);
        \draw [arrow] (concat_op) -- (y);
        \draw [dotted-arrow] (fmap) |- (W3);
        \draw [dotted-arrow] (W3) node[above, xshift=0.9cm] {\tiny $C_n(X^i)$} -| (concat_op);
    \end{tikzpicture}}%
    \caption{(a) is the EnvelopeNet $N_e$ (b) and (c) are networks $N_1$ and $N_2$ obtained by removing filters $C_1$ and $C_2$ respectively.}
    \label{fig:nacanalysis}
\end{figure}


Figure~\ref{fig:Y} shows a single hidden layer of network $N_e$ with $n$ convolution blocks $C_1$, $C_2$....$C_n$, connected in parallel, whose outputs are concatenated. Figure~\ref{fig:Y1} and Figure~\ref{fig:Y2} shows networks $N_1$  and $N_2$ constructed from the original network by pruning convolutional blocks $C_1$ and $C_2$ respectively. In order to retain the dimensionality for doing analysis we use $\phi$, a convolution filter of all zeros with the same dimensions as $C_1$ or $C_2$, as a replacement for the empty cell.

We assume the network $N_e$ to be fully trained and its parameters are inherited by networks $N_1$ and $N_2$. The output of the networks for an input image $X^i$ can be written as:
\begin{equation}
    \begin{split}
        Y^{i}_{e} & = W_{1} X^i\oplus W_{2}X^i\oplus ....W_{n}X^i \nonumber \\
        Y^{i}_{1} & = \phi X^i\oplus W_{2}X^i \oplus ....W_nX^i\nonumber \\
        Y^{i}_{2} & = W_{1}X^i\oplus \phi X^i  \oplus W_{3}X^i \oplus .... W_nX^i \nonumber\\
    \end{split}
\end{equation}

where $W_1$ and $W_2$ are Toeplitz matrices corresponding to the  convolution operations $C_1$ and $C_2$ respectively. 
We define the relative mean square error of the constructed network $N_1$ and $N_2$ as the MSE of the output of the layer relative to the output of the EnvelopeNet $N_e$. 
Without loss of generality, assume that network $N_1$ has higher relative MSE than $N_2$ {\em i.e.} $C_1$ is the correct filter to prune. The relative MSE, between $Y_{1}^{i}$ and $Y_{e}^{i}$ is less than the error between $Y_{2}^{i}$ and $Y_{e}^{i}$.

\begin{equation}
\sum_{i=1}^{m}\|Y_{e}^{i} - Y_{1}^{i}\|_2^2 < \sum_{i=1}^{m}\|Y_{e}^{i} - Y_{2}^{i}\|_2^2\nonumber \end{equation}
\begin{equation}\Rightarrow \sum_{i=1}^{m}\|W_1 X^{i}\|^{2}_{2} < \sum_{i=1}^{m}\|W_2 X^{i}\|^{2}_{2}\nonumber\end{equation}
Here, $m$ is the number of images over which the relative MSE is calculated.
$\sum_{i=1}^{m}\|W_nX^i\|^{2}_{2}$, the sum of the squared $\ell_2$-norm over a given training or validation set, is a featuremap statistic.

This implies that if $N_1$ has lower relative MSE than $N_2$, then the featuremap statistic must be lower for $C_1$, under the assumptions made. 
We can use the squared $\ell_2$ norm as the featuremap statistic to identify the filter in $N$ to prune {\em i.e} to choose between network $N_0$ and $N_1$.
It can be shown that the same featuremap statistic can be be used to choose between any number of filters in parallel in the same layer of a network.
While the analysis is subject to a number of strong assumptions (linear model, single layer) that do not hold in practice, it provides a starting point for the exploration of featuremap statistics.
We conjecture that this metric can be used for more complex networks, and use it in our implementation along with other metrics.
While a real implementation must maximize accuracy not minimize the relative MSE, in practice we find accuracy improves when using this metric as the basis for construction.
Other differences between this analysis and the implemented algorithm include the calculation of the featuremap statistic during training (not after training, as in analysis), and retraining the generated networks rather than using parameter sharing.

\subsection{Construction algorithm}
\begin{algorithm}[]
\caption{Neural Architecture Construction}
\label{alg:constructor}
    \tcp {Neurogenesis: Set network prior}
    $network \leftarrow hyperparams \rightarrow envelopenet$ \\
    \While {$iterations < hyperparams \rightarrow max\_iterations$} {
        \tcp {Learning: Truncated training}
        $filter\_stats \leftarrow short\_train(network)$ \\
        \For{stage in network}{
            $sorted\_filters \leftarrow sort(filter\_stats, stage)$ \\
            \tcp{Apoptosis: Prune sorted filters per stage, subject to constraints: Do not prune a cell if it is the last cell in a layer and limit number of pruned filters per stage}
            $prune\_filters(stage, sorted\_filters)$ \\
            \tcp{Synaptic pruning: Prune skip connections}
            $prune\_skip\_connections(network)$ \\
            \tcp{Hippocampal neurogenesis: Add envelopecell to the tail of stage}
            $network \leftarrow add\_cell(network, stage, hyperparams \rightarrow envelopecell)$\\
        }
        $iterations \leftarrow iterations + 1$ \\
    }
    return $network$
\end{algorithm}
The construction algorithm is shown in Algorithm~\ref{alg:constructor}. It starts by training an EnvelopeNet for $trainingsteps$ steps ({\em neurogenesis})
During the training, statistics from the featuremaps at the outputs of the filters are collected. 
The metric (squared $\ell_2$-norm of the elements of the featuremap of each filter) is calculated over the training set.
After an iteration, the $n_b$ filters with the lowest featuremap statistic within a stage are removed ({\em aptosis}), and an EnvelopeCell is added to the tail of the stage ({\em hippocampal neurogenesis}).
The filters within each stage are then sorted in order of the metric and the filters with the $maxrestructure$ lowest value of the metric are removed, subject to the constraint that every layer must have at least one filter.
Other constraints may be applied, {\em e.g.} $maxrestructure$ may be adjusted based on the number of filters in the layer or the construction may be enabled on a subset of the stages.
An EnvelopeCell is added to the tail end (deepest end) of the stage.
The reason for adding an EnvelopeCell to the end of the stage, is that we do not know, a priori, which filters will improve the performance of the network, so we add an EnvelopeCell, with the understanding that subsequent iterations will remove the unnecessary filters.
We applied DenseNet~\cite{huang16} style skip connections by doing depthwise concatenation on all inputs followed by a 1x1 convolution filter to control the number of output channels.
We assign a scalar weight to all incoming skip connections for every layer. These weights get trained along with the whole network and during the pruning phase we halve the number of skip connections by eliminating connections with lower weight ({\em synaptic pruning}).
The contruction algorithm is run for $N$ iterations.
The result is the network narrows and deepens while maintaining the overall network parameter count approximately same.
The parameters (number of layer per stage, envelope cell, number of stages) and constraints ($maxrestructure$) used for construction are the hyperparameters of the construction algorithm.

The algorithm uses the squared $\ell_2$-norm as the metric. Another metric that was considered was the running feature map variance (filters which have consistently low variance in the distribution of their output featuremap over the training, intuitively contribute less to the classifier's output). 
The feature map variance performed close to, but lower than the squared $\ell_2$ norm.

The construction time is the sum of the time required for the algorithm to run, the time for training to extract the featuremap statistics in each iteration and the time for the training and evaluation of the final network {\em i.e.} $t_{alg}$ + $Nft_{train}$ + $t_{train} + t_{eval}$, where $f$ is the ratio of the time needed to extract featuremap statistics to a full training cycle and $N$ is th number of iterations. For our experiments $f$ was 10 epochs/100 epochs = 0.1 and $N$ was 5.
The algorithm run time and evaluation time are negligible, so total time is $Nft_{train} + t_{train} = (Nf+ 1)t_{train}$ {\em i.e } O($Nf$). 
This compares favorably with evolutionary methods, where the total time is $N_{comb} (t_{train}+ t_{eval})  \approx N_{comb} N_{train}$ where $N_{comb}$ is the number of combinations explored, {\em i.e} O($N_{comb}$)
It also compares well with generative methods where total time is $N(t_{train-rnn} + t_{train} + t_{eval} \approx N(t_{train-rnn} + t_{train})$, {\em i.e.} O($N$) where $N$ is the number of iterations to train the generating network until it generates the final network. 
$f$ represents the reduction in construction time obtained by using the featuremap statistics to reduce training time.



%% file: results.tex
\section{Results}
\label{sec:results}

The algorithm was evaluated on the image classification problem using the CIFAR-10~\cite{krizhevsky09} and ImageNet~\cite{russkovsky15} datasets.
Both construction and the evaluation of the generated networks used a common set of hyperparameters. 
No hyperparameter tuning was done on the EnvelopeNets or the generated networks. 
The training used preprocessing techniques such as random cropping, varying brightness and contrast.
Training used an SGD optimizer with exponentially decaying weight with initial learning rate of $0.1$ and decay factor of $0.1$ per $350$ epochs.
The batch size was set to $50$ for CIFAR-10 and $64$ for Imagenet experiments.
The number of restructuring iterations was 5. 
The number of training steps for the construction algorithm was 10 epochs. 
The number of filters to be pruned in an iteration was 6, subject to a maximum of $1/3$ of the filters in a stage.
Training and evaluation of the base and generated networks ran for at least 100 epochs.
There was no parameter sharing across iterations of the construction algorithm.
The experiments were run using AMLA~\cite{kamath18b} and the results and hyperparameters are available along with the source code~\cite{kamath18c}.

\begin{figure*}
\centering
\includegraphics[width=14cm, height=4cm]{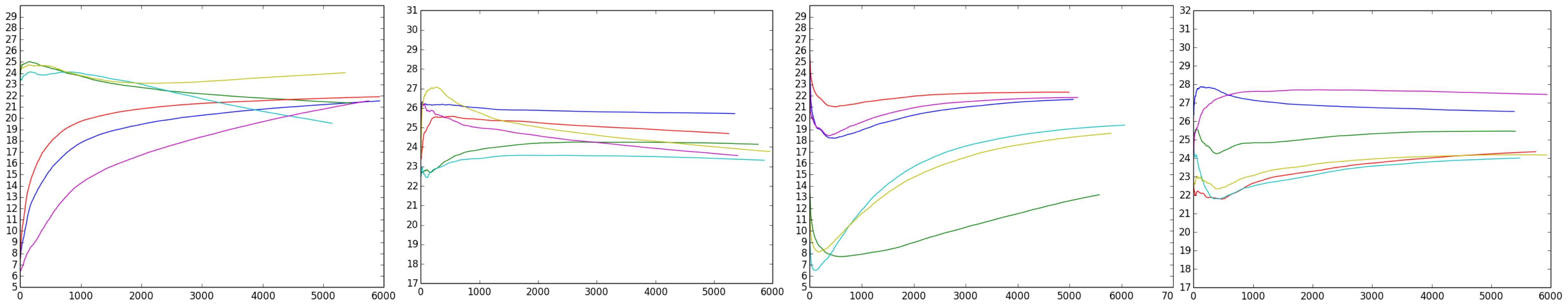}
  \caption{Running featuremap statistic of individual filters at different layers in a network vs. training iterations for a 128/10-1-1-1/6 EnvelopeNet. The graphs show the squared $\ell_2$ norm, normalized by the featuremap size collected over the 10 epochs of training. 
  }
\label{fig:variance}
\end{figure*}

\begin{table}[]
\centering
\begin{tabular}{|C{5.2cm}|C{1.5cm}|C{1.1cm}|C{1cm}|C{1.6cm}|}
    \specialrule{.1em}{.05em}{.05em}
    Network & Dataset & Params & Test error (\%) & Search time (GPU days)\\
    \specialrule{.1em}{.1em}{.1em} 

    \specialrule{.1em}{.05em}{.05em}
    NAS-v3 & CIFAR10 & 37.4M & 3.65 & 1800\\
    \hline
    Block-QNN & CIFAR10 & 39.8M & 3.54 & 96\\
    \hline
    AmoebaNet-B & CIFAR-10 &  34.9M & 2.13 & 3150\\
    \hline
    PNAS & CIFAR10 & \textbf{3.2M} & 3.41 & 225\\
    \hline
    ENAS & CIFAR-10 &  4.6M & 3.54 & 0.45\\
    \hline
    DARTS & CIFAR-10 &  4.6M & 2.76 & 4\\
    \hline
    NAO & CIFAR-10 &  128M & \textbf{2.07} & 200\\
    \hline
    NAC (128/7-6-2/6) & CIFAR10  & 10M & 3.33 & \textbf{0.25}\\
    \specialrule{.1em}{.05em}{.05em}
    NASNet-A & Imagenet & 4.9M & 8.4 & 1800\\
    \hline
    AmoebaNet-A & ImageNet &  6.4M & \textbf{7.6} & 3150\\
    \hline
    DARTS & ImageNet &  \textbf{4.7M} & 8.7 & 4\\
    \hline
    NAC (64/7-6-2/6) & Imagenet  & 9.9M & 11.77 & \textbf{0.25}\\
    \specialrule{.1em}{.05em}{.05em}
    
\end{tabular}
\caption{Accuracy, search time and number of parameters for NAC construction using EnvelopeNets vs. other methods. State of the art numbers are indicated in bold. The NAC experiments were run using single Nvidia V100 GPU. The search time reported here is the sum of the time required for the algorithm to output the final architecture, but does not include the time for the training of the final network. 
}
\label{table:params} 
\end{table}



Figure~\ref{fig:variance} shows the running squared $\ell_2$ norm for filters in different layers of a network vs. training steps. 
After 10 epochs the squared $\ell_2$ norm shows a reasonable separation between each other, although they have not yet converged.
The graphs show that the squared $\ell_2$ norm of feature maps reach  this state within 10 epochs for CIFAR-10 - substantially lower than the number of iterations required to train the network to convergence on the same dataset (usually 100 epochs).
Typically in our experiments $f$, the ratio of the featuremap stability time to convergence time was 0.1, making the construction time 10\% of the time that would be needed, were the candidate networks fully trained to convergence, although the benefit would be less were parameter sharing or early stop used.

\begin{figure*}
\centering
\begin{minipage}[b]{.47\textwidth}
\includegraphics[width=6.5cm,height=5cm]{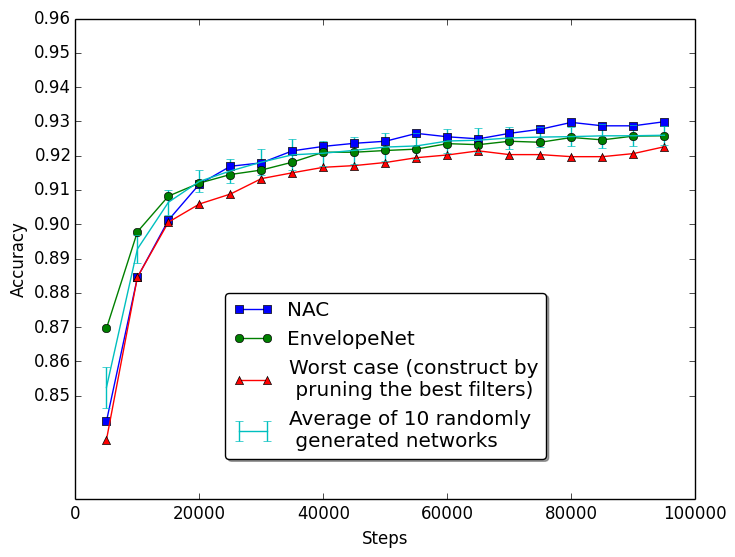}
  \caption{Accuracy vs. training iterations for the EnvelopeNet, the Constructed Net (NAC) and random networks on the CIFAR-10 data set (100 epochs (100K steps))}
\label{fig:perf}
\end{minipage}\qquad
\begin{minipage}[b]{.47\textwidth}
 \includegraphics[width=6.5cm,height=5cm]{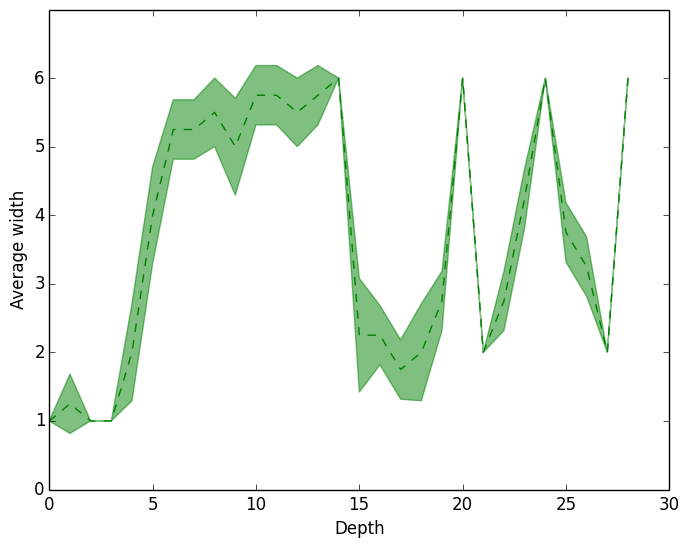}
  \caption{Average width of a layer with standard deviations at different depths for repeated constructions of a 15-6-4-4 network from the same 10-1-1-1 EnvelopeNet}`
\label{fig:structure}
\end{minipage}
\end{figure*}

Figure~\ref{fig:perf} and Table~\ref{table:params} show performance for two sample networks. 

Table~\ref{table:params} shows the performance of a network generated from a 128/2-2-2/6 Envelope with 23.8M parameters, run for 5 iterations to generate a 128/7-6-2/6 Constructed Network for CIFAR10. We use the same network to train on ImageNet and compare along with other methods, with state of the art performance from pther algorithms indicated.

EnvelopeNet A, the Constructed Network A and 10 equivalent random networks (10) were evaluated on the image classification task using CIFAR-10.
The random networks were generated by fixing the depth of the stages equal to the stages in Constructed Network B and adding the same number of blocks, chosen randomly at each stage, subject to a minimum of one block per layer.
Each network was trained on the CIFAR-10 dataset for 100 epochs with performance on the test set evaluated every 5 epochs.
In addition, a worst case network was also constructed using the same construction  algorithm, except it pruned the best performing filters.
Figure~\ref{fig:perf} shows the constructed network clearly outperforming the original EnvelopeNet, the worst case network and the sample average (with standard deviations) of 10 randomly generated networks (it outperformed 9 out of 10 randomly generated networks). 
The performance of the constructed network A is approximately one standard deviation higher than the average random network performance.
Roughly half of the randomly generated networks had lower performance than the EnvelopeNet, indicating that structure is critical for incremental construction - a network with less filters and parameters (the EnvelopeNet) can do better than a network with more filters and parameters.
This indicates that structure of the generated network is responsible for some of the gain, and that the entire gains do not come from deepening the network or increasing the parameters.

Generated Network B from EnvelopeNet B was evaluated on the image classification task using CIFAR 10 and was run for 240 epochs and reached an test error rate of 5.57\%.
Generated Network A (generated using CIFAR-10) was also run on ImageNet. The stem cell for ImageNet was modified by adding 3 layers of convolution filters (3x3, 5x5, 7x7) in parallel to downsample the image from 299x299 to before passing it to the network.
Table~\ref{table:params} shows the number of filters for each layer, the  parameters and flops and accuracy on all datasets for the EnvelopeNets, the generated networks as well as for 2 randomly generated  networks that are equivalent to Constructed Network A.
The construction time in Table~\ref{table:params} is the sum of the time required for the algorithm to run, the time for training to extract the featuremap statistics in each iteration and the time for the training and evaluation of the final network as described in the previous section.

Figure~\ref{fig:structure} shows the average width of the generated network with standard deviation at different depths, for multiple runs of the construction algorithm, constructing different 128/16-5-4-4/6 networks from the same 128/10-1-1-1/6 EnvelopeNet.
Each construction run generates a different sample of featuremap statistics because of the random initialization of the filter weights and preprocessing of the images.
Despite this, the structure of the constructed networks are similar. 
The graph shows that the standard deviation of the width distribution at each layer is in the order of a single filter. 
This indicates that the algorithm tends to prune filters consistently at particular layers within particular stages, which is a strong indication that the algorithm identifies structural improvements.

%% file: discussion.tex
\section{Discussion}
\label{sec:Discussion}
Figure~\ref{fig:perf} indicates that there is a consistent positive difference between the accuracy of the generated vs. the other networks.
The root cause for this may lie in the structure of the network.
Chollet~\cite{chollet16} indicates that different filter types can extract different characteristics.
By providing 3 parallel paths in InceptionNet, with different filters, each inception cell was shown to extract features at different levels.
After training, it was found that for most layers, one of the paths dominated the others, indicating that one path was primarily activated at each layer.
However at the outset it is unclear which paths need activation and which do not, leading a human designer to make a choice, overprovision the network with all possible paths and prune, or do an architecture search with no prior. 
This lead us to the approach of EnvelopeNets and pruning. 

The EnvelopeNet provides a strong prior to the network construction procedure by setting up initial layout of the architecture.
This helps the algorithm restructure the network rather than use resources for the discovery of a base network structure.
The base network structure uses known design practices {\em e.g.}, setting the initial number of layers per stage to reasonable values to limit the parameters, reducing the featuremap's dimension and increasing number of channels periodically after a certain number of layers. 
The prior is provided as a set of hyperparameters to the construction algorithm.

Previous studies have indicated that neural networks exhibit a form of plasticity, allowing random pruning of filters from a network, with little degradation of accuracy provided the fraction of filters pruned is reasonably small~\cite{haoli16}.
However, a counter intuitive result shown recently indicates, that random pruning is as effective as algorithmic pruning when it comes to the accuracy of final trained model~\cite{mittal18}. 
The key difference between our work and these studies is that they are based on  pruning a hand crafted network that has already gone through an optimized design, 
unlike an EnvelopeNet, which is an over provisioned network. 
As we  prune from a large EnvelopeNet to an intermediate  network ({\em e.g.} a handcrafted network) to a fully pruned network, the benefit of pruning may decrease, possibly hitting a knee around the intermediate network.
This would fit in well with our observations as well as results from pruning studies.

Another construction method could be to use reduce a large (deeper/wider) supernet like EnvelopeNet, rather than reduction and addition from a small EnvelopeNet. 
However, the training of a large network increases construction time. Also, the reduction/addition method spreads the restructuring over multiple iterations, rather than in a single iteration, introducing a form of regularization over the architecture search, preventing the structure from overfitting on the artifacts of a single iteration. 
In this regard, the pruning part of the restructuring algorithm, bears some resemblance to the dropout regularization method.
While the motivation behind each method is different, they both remove elements of filters, one probabilistically, during training and the other, physically during construction, although construction permanently removes filters.
The EnvelopeNet construction method lies between bottom up incremental methods of construction and top down reduction methods and provides a reasonable compromise that allows generation of a network without overfitting the structure.

The primary limitation of the restructuring method is that we find the gains from structure appear to reduce when the network parameters increase. In our results we see this when the number of parameters is extremely large, {\em e.g.} if we use a classification block with several fully connected  blocks or as the number of wideners (number of output channels) increases.
Note that in this regime, networks take much larger computation resources to train and any possible gain in the accuracy of network comes from a larger number of parameters rather than intelligent structure design.

%% file: conclusion.tex
\section{Conclusions}

It appears that neural networks exhibit properties that may allow simple heuristic based construction techniques using internal statistics derived during training.
We have exploited these properties in a construction method that can design networks with lower construction time than a search method that evaluates the accuracy of networks.
The generated networks show close to state of the art performance and can outperform most equivalent randomly generated networks, handcrafted networks with equivalent number of parameters and blocks. 
Future directions for work include restructuring algorithms that perform closer to optimal, a deeper understanding of the relationship between structure and performance, and the application of this method to other networks.

%% file: appendix.tex
\appendix
\section{Hyperparameters}

\begin{table}[h]
\centering
\begin{minipage}{\linewidth}
\centering
\begin{tabular}{|C{5.2cm}|C{5.2cm}|}
    \specialrule{.1em}{.05em}{.05em}
    Hyperparameters & Value\\
    \specialrule{.1em}{.05em}{.05em}
    Batch size & 50\\
    \hline
    Optimizer & Momentum, 0.9\\
    \hline
    Learning rate & 0.04\\
    \hline
    Learning rate schedule & Exponential decay per 2 epochs\\
    \hline
    Learning rate decay factor & 0.999\\
    \hline
    Dropout & Keep probability of 0.8\\
    \specialrule{.1em}{.05em}{.05em}
\end{tabular}
\end{minipage}
\caption{Hyperparameters for the candidate networks during construction}
\label{table:hyper-params}
\end{table}

\begin{table}[h]
\centering
\begin{minipage}{\textwidth}
\centering
\begin{tabular}{|C{5.2cm}|C{5.2cm}|}
    \specialrule{.1em}{.05em}{.05em}
    Hyperparameters & Value\\
    \specialrule{.1em}{.05em}{.05em}
    Batch size & 64\\
    \hline
    Optimizer & Momentum, 0.9\\
    \hline
    Learning rate & max - 0.05, min - 0.001\\
    \hline
    Learning rate schedule & Cosine decay per 20 epochs\\
    \hline
    regularization & $\ell_{2}$ norm based, $3\times10^{-4}$\\
    \hline
    gradient clipping & norm based, 5.0\\
    \hline
    Dropout & Keep probability of 0.8\\
    \hline
    Data augmentation & random crop, flip, cutout\\
    \hline
    Stem cell & 3x3 convolution, 128 output channels\\
    \specialrule{.1em}{.05em}{.05em}
\end{tabular}
\end{minipage}\qquad
\caption{Hyperparameters for CIFAR-10 final network}
\label{table:cifar10}
\end{table}
\begin{table}[]
\centering
\begin{minipage}{\textwidth}
\centering
\begin{tabular}{|C{4.cm}|C{6.5cm}|}
    \specialrule{.1em}{.05em}{.05em}
    Hyperparameters & Value\\
    \specialrule{.1em}{.05em}{.05em}
    Batch size & 128\\
    \hline
    Optimizer & Momentum, 0.9\\
    \hline
    Learning rate & 0.1\\
    \hline
    Learning rate schedule & Exponential decay per 25 epochs\\
    \hline
    Learning rate decay factor & 0.97\\
    \hline
    regularization & $\ell_{2}$ norm based, $3\times10^{-5}$\\
    \hline
    gradient clipping & norm based, 5.0\\
    \hline
    Dropout & Keep probability of 0.7\\
    \hline
    Data augmentation & random flip\\
    \hline
    Stem cell & 3x3 convolution, 32 output channels, stride 2\\&3x3 convolution, 64 output channels, stride 2\\&3x3 convolution, 64 output channels, stride 2\\
    \specialrule{.1em}{.05em}{.05em}
\end{tabular}
\end{minipage}
\caption{Hyperparameters for the Imagenet final network}
\label{table:imagenet}
\end{table}
For training of candidate networks we use the  hyperparameters in Table~\ref{table:hyper-params} and for the CIFAR-10 and ImageNet final network we use the hyperparameters in Table~\ref{table:cifar10} and Table~\ref{table:imagenet} respectively. Apart from these hyperparameters, the NAC algorithm requires certain hyperparameters, described below. Some of them are common to most neural architecture search methods and the remaining are specific to our method. The hyperparameters described below are specific to the task of image classification. All hyperparameters are available in configuration files in the code~\cite{kamath18c}.

\subsection{Envelope cell}
 The EnvelopeCell is an overprovisioned cell that uses a set of convolution filters with following kernels: {1x1, 3x3, 3x3sep, 5x5, 5x5sep, 7x7sep} which is a subset of the commonly used kernel search space in architecture search methods.

\subsection{Reduction Cell (Widener)}
The primary purpose of using the reduction cell (also called widener) is to perform downscaling of the image dimensions and increasing the number of channels in the featuremap. We use a simple factorized reduction cell with concatenated output of one 3x3 convolution filter with stride 2 and one max pooling filter with stride 2 connected in parallel. Although many architecture search methods search for the reduction cell, we find that the overall performance is not sensitive to tuning of this reduction cell.

\subsection{Number of stages}
This hyperparameter controls the number of stages in the network. The number of stages should be different depending upon the dataset being used for network design. Each stage is separated by a reduction cell, so there is a limit on number of stage because of the dimension of the input images (most  architecture search methods fix the number of stages by fixing number of reduction cells in their architecture). The networks constructed in this work use 3 stages.

\subsection{Maximum layers per stage}
This hyperparameter  enables construction of a network with a different number of layers in each  stage. It is an array of integers that indicates the maximum number of layers in each stage.  The length of array is equal to the number of stages. Our experiments sets this hyperparameter to [7,6,2].

\subsection{Stage Construction}
 The stage construction hyperparameter allows  construction to be restricted to a subset of all the stages. The hyperparameter is a boolean array with array length being equal to the number of stages with the value indicating whether construction is applied to a particular stage. In our experiments we set this hyperparameter to [true, true, false] for a 3 stage network.

\subsection{Maximum filters to prune}
This hyperparameter limits the number of blocks pruned from each stage. It has a substantial effect on sparsity of the final model obtained and hence total number of parameters in the final model. This hyperparameter is set to 6 in our experiments.

\subsection{Stem Cell}
The stem cell is the initial network which preprocesses the image before it is passed to the constructed network. It is used for increasing the number of channels and downscaling image dimensions to required size. We use a single layer of 3x3 convolution filter with 128 channels as the stem cell during architecture search.

\subsection{Classification Cell}
The classification cell is the final part of the network architecture. It requires one fully connected layer in the end with number of output units equal to the number of labels in the image classification task. We use $4$ layers in our all experiments: the first layer is a global average pooling layer, followed by a flattening layer to convert batch of 3D featuremaps to a batch of vectors. The third layer adds dropout and the fourth layer is the final fully connected layer.